%% file: paper.tex
\documentclass[11pt,a4paper]{article}
\usepackage[hyperref]{acl2020}
\usepackage{times}
\usepackage{latexsym}
\usepackage{hyperref}

\usepackage{url}

\input{Macros/added_packages.tex}

\aclfinalcopy 

\title{Evaluating Machine Common Sense via Cloze Testing}

\author{Ehsan Qasemi\thanks{* Equal Constribution} , Lee Kezar\footnotemark[1], Jay Pujara \and Pedro Szekely \\ 
    Information Sciences Institute,\\
    University of Southern California\\
    Los Angeles, California, USA\\
  {\small \tt \{qasemi, lkezar, jpujara, szekely\}@isi.edu}}

\begin{document}
\maketitle

\begin{abstract}
Language models (LMs) show state of the art performance for common sense (CS) question answering, but whether this ability implies a human-level mastery of CS remains an open question. Understanding the limitations and strengths of LMs can help researchers improve these models, potentially by developing novel ways of integrating external CS knowledge. We devise a series of tests and measurements to systematically quantify their performance on different aspects of CS. We propose the use of cloze testing combined with word embeddings to measure the LM's robustness and confidence. Our results show than although language models tend to achieve human-like accuracy, their confidence is subpar. Future work can leverage this information to build more complex systems, such as an ensemble of symbolic and distributed knowledge.
\end{abstract}

\section{Introduction}
\input{Sections/introduction.tex}

\section{Language Model Evaluation}
\label{sec:LMEval}
\input{Sections/EvaluatingLM.tex}

\subsection{Cloze Test}
\label{sec:cloze}
\input{Sections/cloze.tex}

\subsection{Dispersion} 
\label{sec:clozeDisp}
\input{Sections/ClozeDispersion.tex}

\subsection{Confidence}
\label{sec:confidence}
\input{Sections/confidence.tex}

\section{Experiment 1: Dispersion}
\input{Sections/DispersionExperimentation.tex}

\section{Experiment 2: Confidence}
\input{Sections/ConfidenceExperimentation.tex}

\section{Discussion}

\input{Sections/Discussions.tex}

\bibliography{acl2020}
\bibliographystyle{acl_natbib}

\end{document}

%% file: Macros/added_packages.tex
\usepackage{amsmath,amsfonts,amssymb,amssymb,amsthm}
\usepackage{float}
\usepackage{rotating}  

\usepackage{times}
\usepackage{soul}
\usepackage{url}

\usepackage{colortbl}

\usepackage{subcaption}

\usepackage{array}
\usepackage{hyperref}

\usepackage{xcolor}

\makeatletter

\input{Macros/macros.tex}
\usepackage{subcaption}

%% file: Macros/macros.tex
\renewcommand*{\p@section}{\S\,}
\renewcommand*{\p@subsection}{\S\,}
\renewcommand*{\p@subsubsection}{\S\,}

\definecolor{USCgold}{HTML}{F6C400}
\definecolor{USCred}{HTML}{990000}

\definecolor{myblue}{HTML}{1B587C}
\definecolor{myorange}{HTML}{F6C400}
\definecolor{myred}{HTML}{990000}
\definecolor{mygray}{HTML}{999999}
\definecolor{mygreen}{HTML}{4E8542}
\definecolor{mypurple}{HTML}{604878}

\newcommand{\roberta}{RoBERTa }
\newcommand{\semeval}{SemEval }

%% file: Sections/introduction.tex
Machine common sense (MCS) is a complex skill that rests at the forefront of many NLP tasks, such as reading comprehension, question answering, and generation. It also resists scaling; the task requires a complex representation for many different entities and all the ways they can interact with each other---even atypical ways, as in the canonical example ``An elephant cannot fit through a doorway." Presently, neural language models stand out for their ability to capture meaning in context and apply it in a variety of ways, as indicated by their leading performance on MCS tasks such as aNLI, Social IQA, and Cosmos QA.\footnote{\url{https://leaderboard.allenai.org/}} However, without an empirically-supported theory which links this performance to grounded CS knowledge, improvement is usually reduced to adding more parameters and samples, as in the case of BERT and RoBERTa \citep{liu2019roberta}. Model evaluation should show \textbf{where} and \textbf{to what extent} the model is perplexed on MCS tasks. This information would be instrumental for gathering more representative training data or directing more sophisticated models such as an ensemble approach (i.e. incorporating symbolic knowledge).

In this paper, we utilize \textbf{cloze testing} to enhance our understanding of LM performance on MCS tasks. More specifically, we use pre-trained word embeddings combined with the LM's confidence scores to refine our notions of LM \textbf{accuracy and precision}. This method allows us to locate cases that have neither accuracy nor precision, and more interestingly, cases that have accuracy (e.g., it selects the correct answer on a multiple-choice test), but poor precision (e.g., it was only slightly confident in this choice). 

%% file: Sections/EvaluatingLM.tex
Linguistic common sense is one of the core features for any language model, since such models are learned from massive natural language corpora. Even simple n-gram models for English express CS knowledge such as ``adjectives come before a noun.'' As LMs increase in complexity, they gain the ability to form more precise and accurate representations of lexical semantics in context. As this ability is realized, the LM's CS ability moves beyond linguistic CS and into other categories, such as mathematical, world, and psychological CS.

However, the coverage is far from uniform. State-of-the-art models like RoBERTa-large still produce many nonsensical words such as ``The capital of \textbf{Virginia} is Washington, D.C.'' and ``A \textbf{bamboo} is an animal that eats bamboo.'' (bolded words are the top replacement from RoBERTa). These observations lead us to use the cloze test as an evaluation procedure, which we define in the following subsection.

%% file: Sections/cloze.tex
Cloze testing is the straightforward task of replacing a missing word in a sentence. We use this test for two reasons. First, under certain criteria, the test can exclusively depend on a diverse array of common sense facts. Second, this test is extremely simple to understand and evaluate, especially for LMs which are trained to replace a masked token. We enumerate the test criteria as such:

\begin{enumerate}
    \item[A.] The answer to the blank must be a single word or contraction.
    \item[B.] The answer to the blank should be determinable from the context.
\end{enumerate}

%% file: Sections/ClozeDispersion.tex
Next, we introduce two axioms that will help define MCS in terms of the cloze task:

\begin{enumerate}
    \item[A1:] An LM with accurate common sense knowledge will replace any blank with a set of answers that are semantically similar to the correct answer.
    \item[A2:] An LM with precise common sense knowledge will replace any blank with a set of answers that are semantically similar to each other.
    \item[A3:] Overall common sense should be both accurate and precise.
\end{enumerate}

These axioms will guide our formula for scoring MCS ability. First, to assist with notation, let $\textbf{x}=\{x_1, x_2, \dots x_n\}$ be the context tokens, $x_m$ be the token to mask, and $\textbf{r}_m=\{(w_1, p_1), (w_2, p_2), \dots (w_k, p_k)\}$ be the LM's replacements for $\textbf{r}_m$ and their respective probability.

Intuitively, accuracy can be represented by the average similarity to $x_m$. Although a variety of metrics exist to measure word similarity, we found that cosine similarity via a 50-dimensional GloVe embedding \citep{pennington2014glove} is sufficient. Although some dimensions are heavily skewed, removing them did not cause a significant reduction in accuracy for 65 human-scored word pairs \citep{pawar2018}. Separately, precision can be represented by an average distance to the $\textbf{r}_m$'s mean (i.e. standard deviation).

\begin{align}
    \textrm{acc}(\textbf{r}_m) &= \frac{1}{k} \sum_{i=0}^{k}
    \textrm{cosim}(w_i, x_m) \label{eq:acc} \\
    \textrm{prec}(\textbf{r}_m) &= \frac{1}{k} \sum_{i=0}^{k} \textrm{cosim}(w_i, \textrm{mean}(\textbf{r}_m))\label{eq:prec}
\end{align}

The third axiom can be defined as requiring $\textrm{acc}(\textbf{r}_m) \approx \textrm{prec}(\textbf{r}_m) \approx 1$.

%% file: Sections/confidence.tex
\begin{figure*}[ht]
\centering
\includegraphics[width=\textwidth]{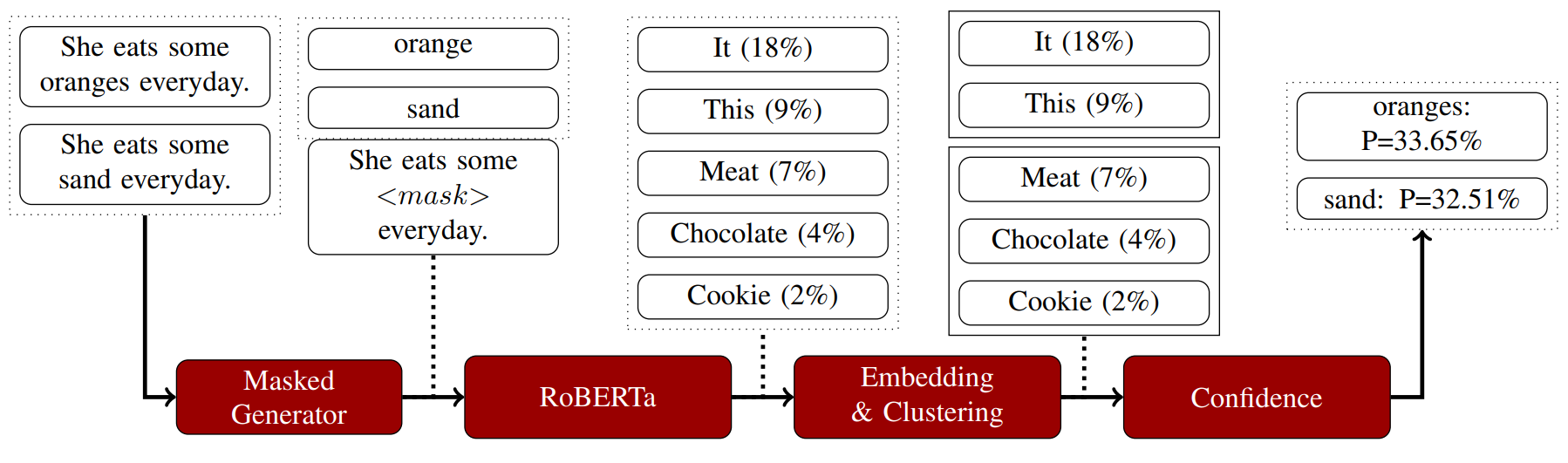}
\caption{Confidence score computation in a nutshell}
\label{fig:confidence}
\end{figure*}

While these metrics leverage distributional semantics to the end of measuring accuracy and precision, they do not incorporate the LM's probability distribution. In response, we introduce a measure of confidence (equation \ref{eq:confidence}) to further refine them. Throughout this section, we focus on the SemEval's task 4A: common sense validation and explanation \citep{semeval2020}, where we are given two sentence that differ in key (set of) words (e.g. \textit{She eats some sand everyday.} vs \textit{She eats some oranges everyday.}). The task is to pick the one sentence that is correct based on the notion of common sense shared by humans and provide an explanation for it; in this case, the correct answer would be the second sentence with an explanation like \textit{oranges are digestible for humans whereas sand is not}.

In order to use \roberta in picking the correct sentence in SemEval, one may use the scores that LM assigns to each sentence and then pick the sentence with higher score. Although this process seems simple, tests on \roberta show that after the fine-tuning, it is capable of picking correct sentence in more than 80\% of the cases \citep{liu2019roberta}.   

In this part, we closely analyze \roberta and investigate its results on \semeval by introducing a notion of confidence (illustrated in figure \ref{fig:confidence}). We note that although we developed and tested the metric on RoBERTa, one can use it for other LMs.

On a high level, we focus on measuring marginal distribution of the choices instead of simply discriminating them with joint distribution of choices. So, instead of having a softmax on the score that LM assigns to each choice (picking the highest score), we compute the probability of each word separately. First, we mask the word that is different in the two choices (\textit{orange}/\textit{sand} in our example) to create a masked form of the sentences that can potentially represent both of them. Then, we ask the LM to propose word candidates that can go into the masked section. In order to find the confidence probability, we cluster the candidate words in a word embedding space and find the distance of the initial word choices (\textit{orange}/\textit{sand}) from the centers of each cluster. Next, we measure the differential distance of the candidate words with the cluster centers as a normalized measure of how each candidate word is close to the clusters; Equation \ref{eq:diff_distance} computes the differential distance for the embedding of word choices $w_i$ from the cluster center $c$.
 \begin{align}
    \label{eq:diff_distance}
    \Delta D(c, w_i) &= 1 - \frac{dist(w_1)}{Z_c} \\
     Z_c &= \sum_{w'}dist(w', c) \nonumber
\end{align}
 Finally, we sum over differential distance of each word choice, multiplied by the overall probability of each cluster to get the confidence probability of the word choice (equation \ref{eq:confidence}). 
\begin{align}
    conf(w) = \sum_c& \Delta D(c, w) * P(c) \label{eq:confidence}
\end{align}
In our setup, the probability of each cluster (equation \ref{eq:prob}) is sum of the probabilities (generate by LM) of all the words in that cluster.
\begin{align}
    P(c) = \sum_{w'' \in c}& P_{w'} \label{eq:prob}
\end{align}

%% file: Sections/DispersionExperimentation.tex
\subsection{Experimental Design}
In Experiment 1, we evaluate RoBERTa's CS ability on 1,000 modified cloze tests. In order to meet the criteria listed in 2.1, we use the training data from SemEval 2020 Task 4A\footnote{\url{http://alt.qcri.org/semeval2020/}}. These sentences are designed to have a very simple syntax, are understandable without a larger context, and do not require specialized knowledge to understand (e.g. \textit{He played the piano with his fingers.}).

We construct approximately $n$ cloze tests for each sentence ($n$ is number of context tokens), where each word (except for stop words and punctuation) is masked. Then, we compute the accuracy and precision as defined in Equations \ref{eq:acc} and \ref{eq:prec}, including extra measures that replace the uniform weight (the $\frac{1}{k}$ factor) with the model's confidence for each word. These four scores are then averaged for each sentence.

\subsection{Results}
Across the corpus, we find that accuracy and precision are moderately correlated with each other ($r=0.390$), and when each term is weighted by the confidence, they are strongly correlated with each other ($r=0.906$). This indicates that we can reasonably predict accuracy from model precision, and vice-versa, but only when the confidence is provided. This makes intuitive sense, as RoBERTa almost always produces inaccurate replacements with low confidence, but if we disregard the confidence score, then they will hold equal weight to much better replacements.
\begin{table}[ht]
    \centering
    \begin{tabular}{c|c}
        sentence & score \\
        \hline
        Grandma knits with thread. & 4.8\% \\
        \hline
        Leopards have spots\\on their bodies. & 17.9\%\\
        \hline
        Dogs shake their tails to\\express their happiness. & 36.4\% \\ 
        \hline
        He parked the car\\in the garage. & 58.9\% \\
        \hline  
        Very few plants and trees\\grow in the desert. & 72.8\% \\
        \hline  
        A passport is necessary to travel\\from one country to another. & 95.0\% \\

    \end{tabular}
    \caption{Some samples from the SemEval training set.}
    \label{tab:sents}
\end{table}

Table \ref{tab:sents} shows some examples of sentences and their scores. As Table \ref{tab:expanded} demonstrates, many of the replacements in low-scoring sentences do not make sense, and this is reflected in the low overall score. Additionally, sentence length is moderately correlated with score ($r=0.548, p<0.0001$), indicating that when there are more ``hints'' for the original word, the task becomes easier for the LM.

\begin{table}[ht]
    \centering
    \begin{tabular}{c|c|c|c}
        Original & Leopards & spots & bodies \\
        \hline
        top-1 & They & scars & backs \\
        top-2 & People & tattoos & back \\
        top-3 & Women & spots & faces \\
        top-4 & Men & scales & heads \\
        top-5 & Children & hair & bodies \\
        Score & 0.09 & 0.20 & 0.25 \\
    \end{tabular}
    \caption{Expanded example for \textit{Leopards have spots on their bodies.}}
    \label{tab:expanded}
\end{table}

%% file: Sections/ConfidenceExperimentation.tex
\subsection{Experimental Design}
We present the empirical results showcasing proposed confidence metric\footnote{Code and data samples available online: \url{https://anonymous-link}} on 100 randomly sampled examples from SemEval-2020. From these 100 samples, we were able to encode 83 examples, in the masked sentence format, meaning the two options in the example did differ on only one n-gram in the fixed sentence position. For measure of distance in equation \ref{eq:diff_distance}, we used the cosine distance implementation in SciPy \citep{scipy} library for simplfy the normalization of distances. Finally, as the word embedding method, we used GloVe \citep{pennington2014glove} pretrained embeddings ($d=50$) and used scikit-learn's \citep{scikit-learn} Gaussian mixture model with $n\_components=2$ for clustering.

\subsection{Results}
\begin{figure}
    \centering
    \includegraphics[width=0.9\linewidth]{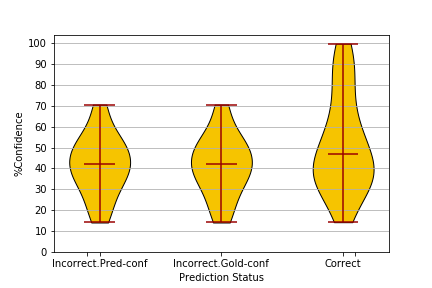}
    \caption{Confidence score of \roberta on a subset of \semeval task 4. From left to right, the violin distributions represent: confidence of predicted label for incorrect cases, confidence of correct label for incorrect cases, and confidence of the predicted label for correct cases. The horizontal bars indicate: min, max, and median of each violin distribution.}
    \label{fig:conf_plot}
\end{figure}

Figure \ref{fig:conf_plot} summarizes the confidence results on the examples. Our results show that the \roberta is able to successfully find the correct answer in 63 of the 83 examples ($\approx$75\% accuracy). However, in both cases (correct and incorrect) its confidence is not high with the mean confidence in both cases less than 50\%. Additionally, our results show a notion of confusion in \roberta in cases that it predicts incorrectly, where the confidence of both options is close.

%% file: Sections/Discussions.tex
These results demonstrate that although state-of-the-art techniques for CS question answering are approaching human-level accuracy, they do not express the human-like confidence or robustness. By definition, CS facts are unambiguous and widely known among humans. Therefore, any model that claims to have common sense should not only be able to pick the most probable answer (as in the SemEval 2020 Task 4A), but also do so with precision and accuracy, i.e. confidence and robustness.
\section{Future Work}
Future work should disentangle the inherent difficulty of a cloze test from the LM's performance via human comparison. In other words, when a LM performs poorly, we can't reasonably attribute it to the model or the sentence itself without knowing the human baseline. This is necessary because many cloze tests are unreasonably difficult, as in the SemEval 2020 stem \textit{The clock is ahead.} 

%% file: paper.bbl
\begin{thebibliography}{6}
\expandafter\ifx\csname natexlab\endcsname\relax\def\natexlab#1{#1}\fi

\bibitem[{Liu et~al.(2019)Liu, Ott, Goyal, Du, Joshi, Chen, Levy, Lewis,
  Zettlemoyer, and Stoyanov}]{liu2019roberta}
Yinhan Liu, Myle Ott, Naman Goyal, Jingfei Du, Mandar Joshi, Danqi Chen, Omer
  Levy, Mike Lewis, Luke Zettlemoyer, and Veselin Stoyanov. 2019.
\newblock Roberta: A robustly optimized bert pretraining approach.
\newblock \emph{arXiv preprint arXiv:1907.11692}.

\bibitem[{Pawar and Mago(2018)}]{pawar2018}
Atish Pawar and Vijay Mago. 2018.
\newblock Calculating the similarity between words and sentences using a
  lexical database and corpus statistics.
\newblock In \emph{IEEE Transactions on Knowledge and Data Engineering}.

\bibitem[{Pedregosa et~al.(2011)Pedregosa, Varoquaux, Gramfort, Michel,
  Thirion, Grisel, Blondel, Prettenhofer, Weiss, Dubourg, Vanderplas, Passos,
  Cournapeau, Brucher, Perrot, and Duchesnay}]{scikit-learn}
F.~Pedregosa, G.~Varoquaux, A.~Gramfort, V.~Michel, B.~Thirion, O.~Grisel,
  M.~Blondel, P.~Prettenhofer, R.~Weiss, V.~Dubourg, J.~Vanderplas, A.~Passos,
  D.~Cournapeau, M.~Brucher, M.~Perrot, and E.~Duchesnay. 2011.
\newblock Scikit-learn: Machine learning in {P}ython.
\newblock \emph{Journal of Machine Learning Research}, 12:2825--2830.

\bibitem[{Pennington et~al.(2014)Pennington, Socher, and
  Manning}]{pennington2014glove}
Jeffrey Pennington, Richard Socher, and Christopher Manning. 2014.
\newblock Glove: Global vectors for word representation.
\newblock In \emph{Proceedings of the 2014 conference on empirical methods in
  natural language processing (EMNLP)}, pages 1532--1543.

\bibitem[{Shirani et~al.(2019)Shirani, Dernoncourt, Asente, Lipka, Kim,
  Echevarria, and Solorio}]{semeval2020}
Amirreza Shirani, Franck Dernoncourt, Paul Asente, Nedim Lipka, Seokhwan Kim,
  Jose Echevarria, and Thamar Solorio. 2019.
\newblock Learning emphasis selection for written text in visual media from
  crowd-sourced label distributions.
\newblock In \emph{Proceedings of the 57th Annual Meeting of the Association
  for Computational Linguistics}, pages 1167--1172.

\bibitem[{{Virtanen} et~al.(2019){Virtanen}, {Gommers}, {Oliphant},
  {Haberland}, {Reddy}, {Cournapeau}, {Burovski}, {Peterson}, {Weckesser},
  {Bright}, {van der Walt}, {Brett}, {Wilson}, {Jarrod Millman}, {Mayorov},
  {Nelson}, {Jones}, {Kern}, {Larson}, {Carey}, {Polat}, {Feng}, {Moore}, {Vand
  erPlas}, {Laxalde}, {Perktold}, {Cimrman}, {Henriksen}, {Quintero}, {Harris},
  {Archibald}, {Ribeiro}, {Pedregosa}, {van Mulbregt}, and
  {Contributors}}]{scipy}
Pauli {Virtanen}, Ralf {Gommers}, Travis~E. {Oliphant}, Matt {Haberland}, Tyler
  {Reddy}, David {Cournapeau}, Evgeni {Burovski}, Pearu {Peterson}, Warren
  {Weckesser}, Jonathan {Bright}, St{\'e}fan~J. {van der Walt}, Matthew
  {Brett}, Joshua {Wilson}, K.~{Jarrod Millman}, Nikolay {Mayorov}, Andrew
  R.~J. {Nelson}, Eric {Jones}, Robert {Kern}, Eric {Larson}, CJ~{Carey},
  {\.I}lhan {Polat}, Yu~{Feng}, Eric~W. {Moore}, Jake {Vand erPlas}, Denis
  {Laxalde}, Josef {Perktold}, Robert {Cimrman}, Ian {Henriksen}, E.~A.
  {Quintero}, Charles~R {Harris}, Anne~M. {Archibald}, Ant{\^o}nio~H.
  {Ribeiro}, Fabian {Pedregosa}, Paul {van Mulbregt}, and SciPy 1.~0
  {Contributors}. 2019.
\newblock \href {http://arxiv.org/abs/1907.10121} {{SciPy 1.0--Fundamental
  Algorithms for Scientific Computing in Python}}.
\newblock \emph{arXiv e-prints}, page arXiv:1907.10121.

\end{thebibliography}
